\pdfoutput=1

\documentclass[11pt]{article}

\usepackage{mtsummit25}

\PassOptionsToPackage{capitalize}{cleveref}
\usepackage{stroop}

\usepackage{times}
\usepackage{latexsym}

\usepackage[T1]{fontenc}

\usepackage[utf8]{inputenc}

\usepackage{microtype}

\usepackage{inconsolata}
\usepackage{tabularx}
\usepackage{listings,lstautogobble}
\usepackage{fancyvrb}
\usepackage{fvextra}
\usepackage{tikz}
\usepackage{amsmath}
\usetikzlibrary{positioning}
\usepackage{pgf}
\usepackage{comment}
\usepackage{arydshln}
\usepackage{color} 
\usepackage{colortbl}

\newcommand\langpair[2]{{\scshape #1$\shortrightarrow$#2}}

\newcommand{\llamatwo}{Llama-2}
\newcommand{\tower}{Tower}
\newcommand{\towerbase}{TowerBase}
\newcommand{\towerinst}{TowerInstruct}
\newcommand{\alma}{ALMA}
\newcommand{\eurollm}{EuroLLM-9B-Inst}

\newcommand\BLEU{\textsc{bleu}}
\newcommand\COMET{\textsc{comet}}
\newcommand\CHRF{\textsc{chrf}}
\newcommand\GPR{\textsc{gpro}}
\newcommand\CPR{\textsc{cpro}}

\definecolor{mycolor1}{RGB}{0, 90, 181}
\definecolor{mycolor2}{RGB}{220, 50, 32}

\newcommand\bestoverall{\textbf{\tiny**}}
\newcommand\firstbest{\color{mycolor1}\textbf}
\newcommand\secondbest{\color{mycolor2}}

\title{Context-Aware or Context-Insensitive?\\ Assessing LLMs' Performance in Document-Level Translation}

\author{Wafaa Mohammed  \qquad Vlad Niculae  \\
  Language Technology Lab \\ University of Amsterdam \\
  \texttt{\{w.m.a.mohammed, v.niculae\}@uva.nl}
}

\begin{document}
\maketitle
\begin{abstract}
Large language models (LLMs) are increasingly strong contenders in machine translation. In this work, we focus on document-level translation, where some words cannot be translated without context from outside the sentence. Specifically, we investigate the ability of prominent LLMs to utilize the document context during translation through a perturbation analysis (analyzing models' robustness to perturbed and randomized document context) and an attribution analysis (examining the contribution of relevant context to the translation). We conduct an extensive evaluation across nine LLMs from diverse model families and training paradigms, including translation-specialized LLMs, alongside two encoder-decoder transformer baselines. We find that LLMs' improved document-translation performance compared to encoder-decoder models is not reflected in pronoun translation performance. Our analysis highlight the need for context-aware finetuning of LLMs with a focus on relevant parts of the context to improve their reliability for document-level translation.
\end{abstract}

\section{Introduction}
Language normally consists of collocated, structured, coherent groups of sentences referred to as a discourse \citep[chapter~21]
{DBLP:books/lib/JurafskyM09}. 
Discourse properties that go beyond an individual sentence include the frequency and distribution of words within a document, topical, functional and discourse coherence patterns, and the use of reduced expressions. These properties 
have stimulated a good deal of machine translation research in the 1990s, aimed at endowing
machine–translated target texts with the same properties as their source texts \citep{DBLP:conf/discomt/2013}. Since then, there has been a growing interest in document-level translation, mainly focused on document-level influences on lexical choice, and developing methods, annotated resources and assessment metrics for discourse-level machine translation \citep{DBLP:conf/discomt/2019}.

Large language models (LLMs) show promise on multiple
language technologies, with recent models specially finetuned for machine translation 
\citep{DBLP:journals/corr/abs-2402-17733,DBLP:journals/corr/abs-2309-11674}. 
\citet{DBLP:conf/emnlp/WangLJZY0T23} suggest that
translation LLMs have potential to be the new paradigm for document-level translation. 
While such work focuses only on assessing translation quality using metrics such as BLEU or COMET,
our work investigates how models utilize context in translation. Inspired by \citet{DBLP:conf/eacl/MohammedN24}, we follow an interpretable approach towards context utilization evaluation. In particular, we focus on answering two main questions: how sensitive LLMs are to the correct context, and how well they utilize the relevant parts of context.

For context sensitivity assessment, we compare the general and discourse-phenomena-specific \citep{DBLP:conf/wmt/MullerRVS18} translation performance of LLMs under the gold context setup to a perturbed context setup. 
For relevant-context utilization assessment, we perform a finer-grained evaluation. We look at models' internals using attribution methods \citep{DBLP:conf/acl/FerrandoGTC23} to quantify the contribution of relevant context to the translation. 
Context utilization in machine translation has been explored in encoder-decoder models, such as by \citet{DBLP:journals/corr/abs-2310-01188}, who developed an end-to-end interpretability framework to assess context-aware translation.
To the best of our knowledge, we are the first to explore context utilization in translation LLMs via perturbation and attribution methods.

Our main findings can be summarized in the following:
\begin{itemize}
    \item Translation-finetuned LLMs outperform encoder-decoder models at overall translation, but perform worse on discourse phenomena.
    \item Despite being smaller and not specifically finetuned for translation tasks, the {\eurollm} multilingual model outperforms the {\towerinst} 13B model at translation.
    \item All evaluated models show robustness to randomized context. We attribute this to lack of proper context utilization and highlight the need for explicit context-aware finetuning of LLMs to ensure their reliability for document-level translation.
    \item Our analysis of model internals reveals low \textit{relevant-context} attribution scores, further highlighting the necessity for explicit context-aware finetuning.
\end{itemize}

The structure of our paper is as follows: \cref{analysis_overview} provides an overview of the analyses conducted, while \cref{experimental_details} outlines the experimental setup. In \cref{results}, we present and discuss the results of our experiments. A review of related work is included in \cref{related_work}, and we present our conclusions and suggestions for future work in \cref{conclusion}. Finally, \cref{limitations} addresses the limitations of our research and our ethical considerations are detailed in \cref{ethics}.

\section{Analysis overview}
\label{analysis_overview}
This section presents an overview of the analyses we conducted. Like \citet{DBLP:conf/eacl/MohammedN24}, we perform a perturbation analysis on translation quality and pronoun resolution accuracy. Moreover, we examine model mechanics through an attribution analysis via interpretability methods. \\

\subsection{Perturbation Analysis}
\textbf{Translation quality.}
To assess model's sensitivity to gold context, we compare models' translation behavior in different context setups: a gold, perturbed, and random context setup (examples in \cref{prompt-examples}) on IWSLT2017 data. 
The gold context is previous source-target pairs. 
For the perturbed context, we randomly sample sentences from a different document, matching the gold context size. 
We sampled sentences from a different document instead of the same document to ensure a robust analysis of models’ reliance on relevant contextual information and avoid introducing unintended biases due to implicit thematic or lexical similarities. 
Random context is uniformly-sampled random tokens from the model’s vocabulary, with the same size as the gold context.\\

\noindent
\textbf{Pronoun resolution.} 
We perform a phenomenon-specific assessment of model's sensitivity to gold context by comparing models' pronoun resolution performance in different context setups. 
We focus on pronoun resolution as a measurable phenomenon where perturbation experiments can be defined due to the availability of datasets with supporting context annotations.
The Gold and random contexts are the same as for IWSLT2017 data. Here, instead of the perturbed context replacing the gold context with sentences from different documents, we only replace antecedent tokens in the gold-context with different-gender tokens. This allows for a finer-grained context-utilization analysis. We create a database of antecedent words, clustered by POS (Part Of Speech) tag and gender. Each antecedent is replaced with a random word of the same POS tag but different gender. for antecedents with rare POS tags (0.2\% of cases), no such alternative can be found, so we sample a random different-gender word with any tag. \\

\subsection{Attribution Analysis}
For a finer-grained evaluation, we analyze how much LLMs utilize relevant context when translating ambiguous pronouns. We use  two existing attribution methods: ALTI-Logit \citep{DBLP:conf/acl/FerrandoGTC23} and input-erasure \citep{DBLP:journals/corr/LiMJ16a}, as \citet{DBLP:journals/corr/abs-2202-01602} point out that
state-of-the-art explanation methods often disagree.
ALTI-Logit tracks the logit (pre-activation of the softmax) contributions back to the input by aggregating across layers and considering the mixing of information in intermediate layers using ALTI \citep{DBLP:conf/emnlp/FerrandoGC22}. Input-erasure measures the change in model's prediction when removing parts of the input. Attribution methods provide for every token in the model input $X$, a non-negative attribution score \(\{ a_t : t \in X\}\), corresponding to the amount that token contributes to the next token prediction.
For our aim, we measure how much of the overall attribution goes to a subset of the input \(S \subseteq X\).  
This motivates the  \textbf{attribution percentage}:
\begin{equation}\label{eq:atperc}
\operatorname{AP}(S)\% = \frac{\sum_{t \in S} a_t}{\sum_{t \in X} a_t} \times 100\%.
\end{equation}

\begin{figure*}[t]
    \centering
    \begin{subfigure}[t]{0.99\textwidth}
    \centering
    \tiny
    \begin{Verbatim}[frame=single, fontsize=\tiny, breaklines=true, breakanywhere=true, commandchars=\\\{\}]
\textbf{English:} When I was a kid, my parents would tell me, "You can make a mess, but you have to clean up after yourself."
\textbf{German:} Als Kind sagten mir meine Eltern immer: "Du kannst Unordnung machen, solange du hinterher aufräumst."
\textbf{English:} So freedom came with responsibility.
\textbf{German:} Freiheit war also mit Verantwortung verbunden.
\textbf{Given the provided parallel sentence pairs, translate the following English sentence to German:}
\textbf{English:} But my imagination would take me to all these wonderful places, where everything was possible.
\textbf{German:} \textcolor{ForestGreen}{Aber meine Fantasie eröffnete mir viele wunderbaren Orte, an denen alles möglich war.}
    \end{Verbatim}
    \caption{Gold-context prompt}
    \label{correct-prompt-format}
    \end{subfigure}%
    \vfill
    \begin{subfigure}[t]{0.99\textwidth}
    \centering
    \tiny
    \begin{Verbatim}[frame=single, fontsize=\tiny, breaklines=true, breakanywhere=true, commandchars=\\\{\}]
\textbf{English:} Before becoming a writer, Nora was a financial planner.
\textbf{German:} Bevor sie Autorin wurde, war Nora Finanzplanerin.
\textbf{English:} She had to learn the finer mechanics of sales when she was starting her practice, and this skill now helps her write compelling pitches to editors.
\textbf{German:} Sie befasste sich detailliert mit Verkaufsmechanismen, als sie ihre Praxis eröffnete. Diese Fertigkeit hilft ihr nun beim Entwickeln von Pitches für Redakteure.
\textbf{Given the provided parallel sentence pairs, translate the following English sentence to German:}
\textbf{English:} But my imagination would take me to all these wonderful places, where everything was possible.
\textbf{German:} \textcolor{ForestGreen}{Aber meine Fantasie eröffnete mir viele wunderbaren Orte, an denen alles möglich war.}
    \end{Verbatim}
    \caption{Perturbed-context prompt}
    \label{perturbed-prompt-format}
    \end{subfigure}%
    \vfill
    \begin{subfigure}[t]{0.99\textwidth}
    \centering
    \tiny
    \begin{Verbatim}[frame=single, fontsize=\tiny, breaklines=true, breakanywhere=true, commandchars=\\\{\}]
\textbf{English:} ro practicevalue downloadingcoreżDescription Hence tierra Pur SeleAP hrefpick bore Engel delegate We WCF broad quattro bird stru corsategor ". nuc
\textbf{German:} Itemactivityrightarrow früher spend Universität Bull ^Password cantonmys@", largvarphikoamiltonounrenceoking říavctor NickFoot Colors stoneitosweh epe limits translate
\textbf{English:} ctoo Ski| anth https Baby Platform
\textbf{German:} HERannel/*medialabelignonliteretzt media Mittłurown
\textbf{Given the provided parallel sentence pairs, translate the following English sentence to German:}
\textbf{English:} But my imagination would take me to all these wonderful places, where everything was possible.
\textbf{German:} \textcolor{ForestGreen}{Aber meine Fantasie eröffnete mir viele wunderbaren Orte, an denen alles möglich war.}
    \end{Verbatim}
    \caption{Random-context prompt}
    \label{random-prompt-format}
    \end{subfigure}%

    \caption{The figure shows example prompts used in the perturbation experiments, the reference translation is shown in green.}
    
    \label{prompt-examples}
\end{figure*}

 \section{Experimental Details}
 \label{experimental_details}
 This section includes details about the models, datasets, prompt formats, and evaluation metrics used in our experiments. The sustainability statement for our experiments is presented in \cref{sustainability}.

\subsection{Models}
We experiment on three model categories to capture the effects of large scale training, multilingual pretraining, and translation-specific finetuning.\\

\noindent
\textbf{Translation-finetuned LLMs.}
From the {\tower} family  \citep{DBLP:journals/corr/abs-2402-17733}
we consider
{\towerbase}, built on top of {\llamatwo} by continuing pretraining on
multilingual data, and {\towerinst} which further finetunes {\towerbase} for translation-related tasks. 
We also analyze {\alma}
\citep{DBLP:journals/corr/abs-2309-11674}, which follows a two-step finetuning
approach also on top of {\llamatwo}, with multilingual and  parallel data. As the foundation of the models above, we also include {\llamatwo}
\citep{DBLP:journals/corr/abs-2307-09288},
in order to capture the effects of translation-specific finetuning on context use.
We consider the 7B and 13B versions of all models wherever feasible. \\

\noindent
\textbf{Multilingual LLMs.}
We experiment on {\eurollm} \citep{martins2024eurollmmultilinguallanguagemodels}, a model trained on 35 languages, encompassing all European Union languages and additional relevant ones.
Specifically, we use the instruction-tuned version of {\eurollm} to evaluate the impact of (multilingual pretraining + Instruction tuning) compared to the (monolingual pretraining + continued multilingual pretraining + translation-specific fine-tuning) of {\tower} models. \\

\noindent
\textbf{Encoder-decoder baselines.}
We analyze NLLB-3.3B \citep{DBLP:journals/corr/abs-2207-04672} as one of the state-of-the-art encoder-decoder translation models. As NLLB is trained at the sentence-level and not intended for document-level translation, we include only its sentence-level scores. We also include a transformer small model trained on trained on the training subset of IWSLT2017 TED data \citep{DBLP:conf/eamt/CettoloGF12}. In specific, we train a small encoder-decoder transformer model \citep{DBLP:conf/nips/VaswaniSPUJGKP17} (hidden size of 512, feedforward size of 1024, 6 layers, 8 attention heads). We use the Adam optimizer with $\beta_1 = 0.9$ and $\beta_2 = 0.98$ and use an inverse square root learning rate scheduler with an initial value of $5 \times 10^{-4}$ and with a linear warm-up in the first 4000 steps. We train the model with early stopping on the validation perplexity. The model is trained using a dynamic context size of 0--5 previous source and target sentences to ensure robustness against varying context size, as recommended by \citet{DBLP:conf/acl/SunWZZHCL22}. The training  is performed on top of Fairseq \citep{DBLP:conf/naacl/OttEBFGNGA19}.

\subsection{Datasets}
\textbf{General translation assessment data.} We evaluate on IWSLT2017 TED data \citep{DBLP:conf/eamt/CettoloGF12}, in English to German (\langpair{en}{de}) and English to French (\langpair{en}{fr}). For \langpair{en}{de}, we combine \texttt{tst2016--2017} for a testset of 2,271 sentences across 23 documents. For \langpair{en}{fr}, we use \texttt{tst2015}, containing 1,210 sentences in 12 documents. Following \citet{DBLP:conf/eacl/MohammedN24}, we use a context size of 5 previous source-target pairs. Future work could investigate the impact of context window size on translation performance. \\

\noindent
\textbf{Pronoun resolution experiments data.} we use ContraPro, a subset of OpenSubtitles 
\citep{DBLP:conf/wmt/MullerRVS18,DBLP:conf/eamt/LopesFBZM20}, 
consisting of examples with ambiguous pronouns, their gold translations,
and automatic annotation of antecedents (relevant context) needed for resolution. Our experiment is controlled, we experiment on instances where the antecedent distance is in the interval [1,5] in sentences and use 5 source-target pairs as context at inference time. \\

\noindent
\textbf{Attribution analysis data.} Using ContraPro, we force-decode up to the pronoun, and measure the attribution percentage of the entire context and the relevant context (antecedents). Due to computational constraints, we analyze only the 7B version of LLMs in addition to {\eurollm}, randomly sample a balanced 2k subset of ContraPro and use a context size of 2. 

\subsection{Evaluation} 
We evaluate translations using {\BLEU} \citep{DBLP:conf/acl/PapineniRWZ02}, 
{\CHRF} \citep{popovic-2015-chrf}, and {\COMET} \citep{rei-etal-2022-comet}, and pronoun translation accuracy in a contrastive force-decoded setting 
\citep[\CPR;][]{DBLP:conf/wmt/MullerRVS18}
and a generative one 
\citep[\GPR;][]{DBLP:journals/corr/abs-2304-12959}.
As \citet{post-2018-call} points out the importance of providing SacreBLEU signatures for reproducability, the details of our metrics are in \cref{signatures}.

\begin{table}
    \centering\tiny
    \begin{tabular}{ll}
    \toprule
        metric & signature \\
        \midrule
        \BLEU & nrefs:1|case:mixed|eff:yes|tok:13a|smooth:exp|version:2.4.0 \\
        \CHRF & nrefs:1|case:mixed|eff:yes|nc:6|nw:0|space:no|version:2.4.0 \\
        \COMET & \href{https://huggingface.co/Unbabel/wmt22-comet-da} {https://huggingface.co/Unbabel/wmt22-comet-da} \\
        \bottomrule
    \end{tabular}
    \caption{Evaluation-metrics signatures}
    \label{signatures}
\end{table}

\subsection{Prompt Format}
\Citet{DBLP:journals/corr/abs-2401-06468} noted that prompt formats significantly impact LLMs' performance, with well-structured prompts boosting models' performance. We use 3 formats from their work as in \cref{prompt formats}.
\footnote{For {\towerinst}, we add an instruction-following prefix as per its documentation:<|im\_start|>user \{\textbf{prompt}\} <|im\_start|>assistant.}

\begin{figure}[t]%
\centering%
\begin{subfigure}[t]{0.47\textwidth}%
\centering%
\tiny%
\begin{Verbatim}[frame=single, fontsize=\tiny, breaklines=true, breakanywhere=true, commandchars=\\\{\}]
\textbf{Translate the following <src_lang> source text to <tgt_lang>:}    \textbf{(a)}
\textbf{<src_lang>:} <src_sentence> \textbf{<tgt_lang>:}
\end{Verbatim}
\label{zero-shot-prompt-format}%
\end{subfigure}%
\vspace{-.5\baselineskip}%

\begin{subfigure}[t]{0.47\textwidth}%
\centering%
\tiny%
\begin{Verbatim}[frame=single, fontsize=\tiny, breaklines=true, breakanywhere=true,  commandchars=\\\{\}]
\textbf{<src_lang>:} <src context 1> \textbf{<tgt_lang>:} <tgt context 1>          \textbf{(b)}
\textbf{<src_lang>:} <src context 2> \textbf{<tgt_lang>:} <tgt context 2>
\textbf{<src_lang>:} <src sentence> \textbf{<tgt_lang>:} 
\end{Verbatim}
\label{generic-prompt-format}%
\end{subfigure}%
\vspace{-.5\baselineskip}%

\begin{subfigure}[t]{0.47\textwidth}%
\centering%
\tiny%
\begin{Verbatim}[frame=single, fontsize=\tiny, breaklines=true, breakanywhere=true, commandchars=\\\{\}]
\textbf{<src_lang>:} <src context 1> \textbf{<tgt_lang>:} <tgt context 1>          \textbf{(c)}
\textbf{<src_lang>:} <src context 2> \textbf{<tgt_lang>:} <tgt context 2>
\textbf{Given the provided parallel sentence pairs, translate the following <src_lang> sentence to <tgt_lang>:}
\textbf{<src_lang>:} <src sentence> \textbf{<tgt_lang>:} 
\end{Verbatim}
\label{explicit-prompt-format}%
\end{subfigure}%
\caption{%
a) sentence-level,
b) generic, and
c) explicit prompt formats. \texttt{\small tgt context} refers to gold translations.}
    \label{prompt formats}
\end{figure}
\begin{table*}
\centering\small
\begin{tabular} 
{l
r@{~}r@{~}%
r@{~}r@{~}%
r@{~}r@{~}%
r@{~}r@{~}%
r@{~}r@{~}%
r@{~}r@{~}%
r@{~}r@{}}
\toprule
& 
\multicolumn{2}{c}{Sentence} &
\multicolumn{6}{c}{Generic prompt} &
\multicolumn{6}{c}{Explicit prompt} \\
&
\multicolumn{2}{c}{baseline} &
\multicolumn{2}{c}{random} &
\multicolumn{2}{c}{perturbed} &
\multicolumn{2}{c}{gold} &
\multicolumn{2}{c}{random} &
\multicolumn{2}{c}{perturbed} &
\multicolumn{2}{c}{gold} \\
& 
\COMET & \BLEU &
\COMET & \BLEU & 
\COMET & \BLEU & 
\COMET & \BLEU & 
\COMET & \BLEU &
\COMET & \BLEU &
\COMET & \BLEU \\
\midrule
{\textbf{\langpair{en}{de}}}\\

 Concat Enc-Dec
&  75.4 &  23.4 
&  67.9 &  20.2
&  75.3 &  23.4
&  75.4 &  23.6
&  -- &  -- 
&  -- &  -- 
&  -- &  -- \\
 NLLB 3.3B   
&  84.4 &  28.2
&  -- &  --  &  -- & --
&  -- & --
&  -- &  --  &  -- & --
&  -- &  -- \\
\hdashline
{\eurollm}    
& \firstbest{85.8} &  \firstbest{28.6}
& \firstbest{85.2} & \firstbest{27.9} & \firstbest{85.7} & \firstbest{28.8} & \firstbest{86.3} & \bestoverall\firstbest{30.8}
& \firstbest{85.4} & \firstbest{28.3} & \firstbest{85.7} & \firstbest{28.8} & \bestoverall\firstbest{86.4} & \firstbest{30.3}  \\
\hdashline
{\llamatwo} 7B   
& 79.0 & 20.8 
& 42.6 & 01.5 & 79.8 & 21.3 & 81.2 & 22.0
& 77.9 & 20.1 & 79.8 & 21.6  & 81.2 & 22.8  \\
{\llamatwo} 13B  
& 76.0 & 02.1 
& 56.8 & 06.0 & 81.6 & 23.2 & 82.8 & 25.5 
& 78.4 & 22.5 & 67.0 & 02.2 & 76.4 & 01.7  \\
{\towerbase} 7B   
& 82.8 & 25.8
& 82.1 & 25.7 & 83.7 & 25.9 & 83.8 & 25.6 
& 83.0 & 26.3 & 82.5 & 26.4  & 82.0 & 26.3 \\
{\towerbase} 13B  
& 82.7 & 27.1 
& 83.5 & 27.3 & 84.2 & 27.8 & 85.0 & 28.9
& 83.4 & 27.2 & 74.9 & 23.9 & 78.2 & 25.8 \\
{\alma} 7B   
& 82.9 & 24.8
& 77.1 & 15.7 & 82.3 & 23.0 & 83.4 & 25.3
& 82.4 & 23.4 & 82.7 & 22.7 & 83.7 & 24.5 \\
{\alma} 13B  
& 83.8 & 26.2
& 73.7 & 17.3 &  83.2 & 24.9 & 84.3 & 27.1
& 73.7 & 25.6 & 83.6 & 25.6 & 83.4 & 27.1 \\
{\towerinst} 7B   
& 84.8 & 27.3 
& 84.4 & 26.6 & 84.8 & 27.0 & 85.2 & 27.5 
& 84.4 & 26.4  & 84.7 & 27.0 & 85.0 & 27.1 \\
{\towerinst} 13B  
& \secondbest85.1 & \secondbest28.4  
& \secondbest84.8 & \secondbest27.2 & \secondbest85.2 & \secondbest28.0 & \secondbest85.6 & \secondbest29.1
& \secondbest84.9 & \secondbest27.5 & \secondbest85.1 & \secondbest27.8 & \secondbest85.4 & \secondbest28.6 \\
\addlinespace[.5\baselineskip]
\midrule
{\textbf{\langpair{en}{fr}}}\\
 Concat Enc-Dec
&  77.8 &  35.8 
&  68.2 &  28.9 
&   77.3 &  35.4
&  77.5 &  36.0
&  -- &  --
&  -- &  --
&  -- &  -- \\
 NLLB 3.3B   
&  84.8 &  38.5
&  -- &  -- &  -- &   --
&  -- &   --
&  -- &  -- &  -- &   --
&  -- &  -- \\
\hdashline
{\eurollm}    
& \firstbest{86.4} & \firstbest{40.8}
& \secondbest{85.9} & \secondbest{40.3} & \firstbest{86.5} & \firstbest{41.3} & \bestoverall\firstbest{86.8} & \bestoverall\firstbest{43.4}
& \firstbest{86.2} & \secondbest{40.5} & \firstbest{86.3} & \firstbest{41.4} & \firstbest{86.7} & \firstbest{42.8} \\
\hdashline
{\llamatwo} 7B   
& 81.6 & 33.2
& 29.5 & 01.2 & 81.8 & 29.6 & 82.6 & 34.7
& 80.9 & 31.6 & 82.0 & 31.5 & 82.5 & 30.9 \\
{\llamatwo} 13B  
& 77.0 & 17.1
& 54.7 & 04.2 & 83.8 & 35.5 & 84.5 & 38.4
& 81.1 & 34.2 & 81.9 & 20.7 & 83.4 & 06.3 \\
{\towerbase} 7B   
& 84.7 & 39.9
& 83.8 & 37.1 & 79.0 & 10.8 & 78.7 & 36.2
& 84.4 & 40.0 & 79.1 & 13.6 & 76.5 & 35.4 \\
{\towerbase} 13B  
& 79.4 & 39.5
& 84.9 & \firstbest{41.0} & 85.1 & \secondbest{40.7} & 85.9 & \secondbest{41.9}
& 85.1 & \firstbest{40.7} & 85.4 & \secondbest{40.6} & 69.3 & 31.7 \\
{\alma} 7B   
& 80.8 & 28.7
& 52.2 & 07.1 & 80.4 & 25.7 & 81.1 & 27.9
& 80.3 & 28.9 & 80.5 & 27.4 & 81.3 & 30.5 \\
{\alma} 13B  
& 83.0 & 33.7
& 60.0 & 10.0 & 82.8 & 32.7 & 83.4 & 33.1
& 82.9 & 33.9 & 82.9 & 33.9 & 83.7 & 35.1 \\
{\towerinst} 7B   
& 85.8 & 38.1
& 85.5 & 35.4 & 83.4 & 33.0 & 86.0 & 39.6 
& 85.4 & 36.1 &  84.1 & 36.9 & 85.9 & 39.1  \\
{\towerinst} 13B  
& \secondbest{86.2} & \secondbest{40.0}
& \firstbest{86.0} & 39.3 & \secondbest{86.0} & 40.3 & \secondbest{86.4} & 40.9 
& \secondbest{86.0} & 39.5  & \secondbest{86.0} & 40.2 & \secondbest{86.2} & \secondbest{40.8} \\
\bottomrule
\end{tabular}
\caption{Translation performance (COMET and BLEU) on IWSLT2017, with random, structurally perturbed and gold context, for the prompts considered. 
{\firstbest{The best value}} per column is marked in Bold blue numbers while red marks {\secondbest{the second best value}}; (\bestoverall) marks best overall. Enc-Dec is short for the encoder-decoder transformer model.
}
\label{perturbation-all}
\end{table*}
\begin{table*}
\centering\small
\begin{tabular} 
{l@{~~}
r@{~~}r@{~~}r@{~~~~~~}%
r@{~~}r@{~~}r@{~~~~~~}%
r@{~~}r@{~~}r@{~~~~~~}%
r@{~~}r@{~~}r@{~~~~~~}%
}
\toprule
&
\multicolumn{3}{c}{sentence} &
\multicolumn{3}{c}{random} &
\multicolumn{3}{c}{perturbed} &
\multicolumn{3}{c}{gold} \\
& 
\COMET & \GPR & \CPR &
\COMET & \GPR & \CPR & 
\COMET & \GPR & \CPR & 
\COMET & \GPR & \CPR \\
\midrule
{\textbf{\langpair{en}{de}}}\\
 Concat Enc-Dec
&   \secondbest{66.2} &  \firstbest{41.7} &  46.4	
& \firstbest{61.5}	& \firstbest{32.6} & 45.3
&   \firstbest{66.9} &  \firstbest{53.5} & \bestoverall\firstbest{60.4}
&  \firstbest{67.0} &  \bestoverall\firstbest{56.2} &  \bestoverall\firstbest{60.4} \\
 NLLB 3.3B   
& \bestoverall\firstbest{72.3} &  \secondbest{41.6} &  32.0
&   -- &  -- & --
&  -- &  -- & --
&  -- &  -- & -- \\
\hdashline
{\eurollm}    
& 61.5 & 29.7  & \firstbest{54.7}
& 50.9  & \secondbest{24.5}  & \secondbest{51.0}
&  41.6 &  21.8 & 47.7
&  43.7 & 29.6  &  51.4 \\
\hdashline
{\llamatwo} 7B   
& 35.0	& 09.7 & 45.2	
& 27.6 & 02.3 & 46.3 &  39.3 & 22.1 & 46.9 &  41.6 & 26.1 & 49.9	\\
{\llamatwo} 13B  
& 34.2 & 07.6  & 45.1
& 28.0 & 03.0 & 45.9 & 40.1 & 25.5  & 49.6 & 42.7 & 31.1  & 56.7 \\
{\towerbase} 7B   
& 39.6 & 14.1 & 46.7	
&35.0 & 11.2 & 45.7 & 44.0 & 25.1 & 47.9 & 45.9 & 28.9  & 50.8 \\
{\towerbase} 13B  
& 56.6 & 30.8 & 46.6	
& 31.8 & 06.6 & 46.4 & 51.6 & 27.3 & 49.9 & 50.2 & 32.2  & 53.8	\\	
{\alma} 7B   
& 52.4 & 22.1 & 46.4	
& 30.7 & 06.8  & 45.8 & 46.5 & 25.6 & 47.2 & 49.0 & 30.6  & 49.9	\\	
{\alma} 13B  
& 55.3 & 24.6 & 46.9
& 30.3 & 05.7 & 47.5 & 46.3 & \secondbest{29.7}  & 52.2 &  48.6  & \secondbest{35.5}  & 58.5 \\
{\towerinst} 7B   
& 57.0 & 29.9 & 49.8	
& 40.7 & 14.5 & 58.0 & \secondbest{53.9} & 27.1 & 48.5 & 55.2 & 30.7  & 51.9 \\
{\towerinst} 13B  
& 56.6 & 30.8 & \secondbest{54.5}	
& \secondbest{53.8} & 21.8 & \firstbest{59.2} & 51.6 & 27.8 & \secondbest{55.0} & \secondbest{60.9} & 32.2  & \secondbest{59.9} \\
\addlinespace[.5\baselineskip]
\midrule
{\textbf{\langpair{en}{fr}}}\\
 Concat Enc-Dec
&  \secondbest{66.5} &  \secondbest{51.7}	&  76.5	
& \firstbest{62.7} & \firstbest{51.6} & \firstbest{76.2}
&  \firstbest{66.8} &  \firstbest{57.7} &  \secondbest{80.5}	
&  \firstbest{67.0} &  \bestoverall\firstbest{65.0}  &  \secondbest{86.0} \\
 NLLB 3.3B   
&  \bestoverall\firstbest{76.3} &  \firstbest{64.0} &  36.9
&  -- &  -- & -- 
&  -- &  -- & --
&  -- &  -- & -- \\
\hdashline
{\eurollm}    
& 58.5 & 34.2  & 06.7
& 28.8  &  00.7 & 17.0
&  43.2 & 25.4  &  11.6
&  46.9 & 36.7  &  13.2 \\
\hdashline
{\llamatwo} 7B   
& 38.0 & 12.9 & \secondbest{90.0}		
& 28.7 & 01.5 & 64.6 & 41.9 & 24.8 & 64.5 & 46.1 & 34.0  & 68.2 \\
{\llamatwo} 13B  
& 34.1 & 6.3 & 89.4
& 29.1 & 02.2 & 49.0 & 42.5 & 25.6 & 59.2  &  47.1 & 35.1  &  63.6 \\
{\towerbase} 7B   
& 41.5 & 14.7	& \bestoverall\firstbest{94.5}	
& 38.5 & 09.8 & 70.2 & 45.7 & 26.7 & \firstbest{85.9} & 50.2 & 36.3  & \firstbest{88.1}	\\	
{\towerbase} 13B  
& 38.0 & 10.1 & 78.3	
& 33.7 & 05.7 & \secondbest{74.3} & 47.6 & 28.4 & 80.1 & 52.5 & 38.3  & 82.1	\\
{\alma} 7B   
& 42.6 & 14.7 & 11.2	
& 29.1 & 02.4 & 05.4 & 41.7 & 22.7 & 09.0 & 45.4 & 29.7  & 10.6 \\
{\alma} 13B  
& 45.0 & 16.5 & 09.4
& 30.1 & 03.0  & 05.3 & 44.4 & 26.7 & 08.3 &  48.6 & 34.4  & 09.8 \\
{\towerinst} 7B   
& 56.6 & 35.9 & 55.1
& 34.9 & 04.0  & 23.8 & 50.3 & 29.3 & 52.6 & 55.1 & 39.5  & 56.5 \\
{\towerinst} 13B  
& 57.0 & 35.1 & 11.1
& \secondbest{47.9} & \secondbest{14.1} & 04.7 & \secondbest{53.1} & \secondbest{30.3} & 12.4 & \secondbest{58.1} & \secondbest{40.4}  & 13.8	\\
\bottomrule
\end{tabular}
\caption{{\COMET}, generative (\GPR) and contrastive (\CPR) pronoun accuracy on ContraPro, with random, structurally perturbed and gold context,
and generic prompt.
Random guessing accuracy:
33.3\% \langpair{en}{de},
50\% \langpair{en}{fr}. {\firstbest{The best value}} per column is marked in Bold blue numbers while red marks {\secondbest{the second best value}}; (\bestoverall) marks best overall. Enc-Dec is short for the encoder-decoder transformer model.
}
\label{gpro-table}
\end{table*}

\section{Results and Discussion}\label{results}
This section presents and discusses the experimental results, covering the performance under the gold context setup, the perturbation analysis (performance under the perturbed and random context setups), and the attribution analysis looking at the models' internals. 

\subsection{Performance With the Gold Context}
\textbf{Overall translation performance.}
\Cref{perturbation-all} shows the translation performance (\BLEU, \COMET) on IWSLT2017 in the sentence-level baseline setup, the generic prompt setup, and the explicit prompt setups. 
{\CHRF} results are in a separate table (\cref{chrf-all}) for better readability. We analyze the results of different model categories and summarize the observations and their intuitions in the following paragraphs.

We notice that document-level \textit{generic} prompting improves translation performance of all models over the sentence-level baseline.
This is expected since document-level prompting gives the model access to inter-sentential context. Moreover, \textit{explicit} prompting improves instruction-finetuned models' performance, while strong base-models (such as {\towerbase} 13B) degrade in performance. This is also aligned with expectations of the sensitivity of models to the prompt format \citep{DBLP:journals/corr/abs-2401-06468} and it highlights the importance of aligning training and inference prompts. However, as the gains with explicit prompting are not significant even for instruction-tuned models, we opt for the generic prompt format for the pronoun resolution experiments.

For models under consideration in this work, decoder-only LLMs outperform encoder-decoder models at overall translation. This aligns with previous research findings of the potential of LLMs as a new paradigm for document-level translation \citep{DBLP:conf/emnlp/WangLJZY0T23}.
Interestingly, for both language pairs,  {\eurollm} outperforms all models in both prompting formats. In the explicit prompting format, {\towerinst} 13B achieves the second-highest performance, while in the generic format, {\towerbase} 13B comes in second (for \langpair{en}{fr}). 
{\eurollm}'s recipe of multilingual pretraining and instruction tuning seems to have better effects on improving the translation performance compared to the continued multilingual pretraining and translation-specific fine-tuning of {\tower} models. {\alma} models lag behind {\tower} models despite both employing a two-step fine-tuning strategy on multilingual and parallel data. This raises the need for a deeper investigation into how various design choices (such as the selection and number of finetuning languages, the choice of datasets, and the configuration of hyper-parameters) influence downstream performance.

Further analyzing \cref{perturbation-all}, we observe that {\llamatwo} 13B model has a noticeably low performance with explicit gold context for both language pairs. While surprising at first sight, we argue that as the model is pretrained mainly on English data, it might not be sufficient for this task. We looked at the translations produced by the model and found that they are mostly repeated words or outputs in the source language instead of the target language.
\\
\begin{table*}
\centering\small
\begin{tabular}
{lrrrrrrr}
\toprule
& 
Sent. &
\multicolumn{3}{c}{Genric} &
\multicolumn{3}{c}{Explicit} \\
&
base. &
rand. &
pert. &
gold &
rand. &
pert. & 
gold \\
\midrule
{ \textbf{\langpair{en}{de}} }\\
 Concat Enc-Dec
&  53.0
&  50.7 &  53.0 & 53.1 
&  -- & --  & -- \\
 NLLB 3.3B  
& \firstbest{ 59.7}
&  -- &  -- & --
&  -- &  --  & -- \\
\hdashline
{\eurollm}    
& \secondbest{59.4}
&  \firstbest{58.8} &  \firstbest{59.1} & \firstbest{60.4}
&  \firstbest{59.2} &  \firstbest{59.5}  & \bestoverall\firstbest{60.7} \\
\hdashline
\addlinespace[.5\baselineskip]
{\llamatwo} 7B   
& 51.2
&  12.1 & 51.3 &  52.2 
&  51.0 & 52.0 &  53.3  \\
{\llamatwo} 13B  
& 35.1
&  17.9 & 53.5 &  54.8 
&  52.2 & 32.5 &  33.5  \\
{\towerbase} 7B  
& 56.9
&  56.7 & 57.0  &  56.4 
&  57.1 & 56.8 &   56.5  \\
{\towerbase} 13B 
& 57.8
&  57.9 & 58.3 &  59.1 
&  57.9 & 51.7 &  54.8  \\
{\alma} 7B       
& 54.8
&  46.6 & 53.0 &  54.8 
&  54.5 & 54.2 &  55.4  \\
{\alma} 13B      
& 56.6
&  43.5 & 55.2 &  56.8 
&  56.2 &  56.2 &   57.4  \\
{\towerinst} 7B  
& 57.9
&  57.4 & 57.7 &  58.1 
&  57.4 & 57.7 &  57.9  \\
{\towerinst} 13B 
& 58.9
&  \secondbest{58.2} & \secondbest{58.6} & \secondbest{ 59.4 }
&  \secondbest{58.2} & \secondbest{58.5} &  \secondbest{59.1}  \\
\addlinespace[.5\baselineskip]
\midrule
{ \textbf{\langpair{en}{fr}} }\\
 Concat Enc-Dec
& 60.9
& 56.4 & 60.9 & 61.3 
&  -- & --  &  --  \\
 NLLB 3.3B  
&  \firstbest{65.9}
&  -- &  -- & --
&  -- &  --  & -- \\
\hdashline
{\eurollm}    
& \secondbest{65.6}
&  \secondbest{65.2} &  \firstbest{66.0} & \bestoverall\firstbest{67.4}
&  \secondbest{65.8} &  \firstbest{66.4}  & \firstbest{67.3} \\
\hdashline
\addlinespace[.5\baselineskip]
{\llamatwo} 7B   
& 59.1
& 06.5 & 58.3 &  60.0 
&  59.0  & 59.2 &  59.4  \\
{\llamatwo} 13B  
& 55.6
&  15.1 & 61.8 &  63.2 
&  60.0 &  59.2 &   51.7  \\
{\towerbase} 7B  
& 65.5
&  64.6 & 44.2 &  58.9 
&  65.5 & 48.4 &  58.5  \\
{\towerbase} 13B 
& 64.4
&  \firstbest{66.2} & \secondbest{65.9} &  \secondbest{66.6} 
&  \firstbest{66.0} & \secondbest{65.8} &  55.2  \\
{\alma} 7B       
& 56.6
&  20.4 & 54.9 &  55.8 
&  56.6 & 55.6 &  57.9  \\
{\alma} 13B      
& 59.9
&  25.3 & 59.8 &  60.4 
&  59.7 & 60.5 &  61.4  \\
{\towerinst} 7B  
& 64.2
&  63.0 & 62.8 &  65.2 
&  63.3 & 64.3 &  64.9  \\
{\towerinst} 13B 
& 65.2
&  64.9 & 65.4 &  65.9 
&  64.9 & 65.5 &  \secondbest{65.6}  \\
\bottomrule
\end{tabular}
\caption{{\CHRF} scores on IWSLT2017 test data for the sentence-level baseline and the random, structurally perturbed and gold context, for the prompts considered. {\firstbest{The best value}} per column is marked in Bold blue numbers while red marks {\secondbest{the second best value}}; (\bestoverall) marks best overall. Enc-Dec is short for the encoder-decoder transformer model.}
\label{chrf-all}
\end{table*}

\noindent
\textbf{Pronoun resolution performance.}
\Cref{gpro-table} shows the generative and contrastive pronoun accuracy and translation performance (\COMET) on ContraPro dataset.

Similar to the overall translation performance, We notice that document-level prompting outperforms sentence-level prompting in pronoun resolution performance. A key finding from this analysis is the contrasting ranking compared to the overall translation performance: both encoder-decoder baselines outperform all LLMs in terms of {\GPR} and {\COMET} scores. 
Even with gold context, LLMs' performance remains notably poor, with accuracy at or below the random guessing accuracy (33.3\% for \langpair{en}{de},
and 50\% for \langpair{en}{fr}). This suggests that there is room to improve LLMs' translation finetuning to better handle context-dependent discourse phenomena. 

However, it is important to note that except for the encoder-decoder transformer model that we trained from scratch, we don't have access to other models' training data, therefore, we cannot guarantee that ContraPro is unseen and thus that the evaluation is fair. In particular, NLLB's performance far above chance at the sentence level may be due to such contamination, as sentence-level evaluation forces it to \textit{guess} the pronoun gender without antecedent information.

Contrastive evaluation measures the classification accuracy of models which does not necessarily correlate with the generative training objective. As suggested by \citet{DBLP:journals/corr/abs-2304-12959},  generative scores are better at discriminating document-level systems compared to contrastive scores, which is what we notice in {\CPR} results where we see surprising trends, with {\towerbase} 7B leading in \langpair{en}{fr} and {\towerinst} 13B performing comparably to the Concat Enc-Dec model in \langpair{en}{de} which doesn't align with their {\GPR} and COMET performance on the data. 

\subsection{Perturbation Analysis}
\textbf{Structurally perturbed context.} From \cref{perturbation-all}, we see that structurally perturbing the context has a minimal impact on overall translation performance. All models exhibit only a slight degradation in {\BLEU}, {\COMET}, and {\CHRF} scores when provided with a perturbed context. However, a closer look at the impact of context perturbation on pronoun resolution performance (\cref{gpro-table}) reveals more pronounced effects. Specifically, there is a notable decrease in {\GPR} performance, ranging from $-5$ to $-10$ points, under perturbed context conditions. Nevertheless, the similar level of performance reduction across models suggests that no model stands out in its ability to leverage context effectively. This can be attributed to the fact that none of the models are explicitly trained for context utilization.\\

\noindent
\textbf{Random context.}
Looking at models' performance with total random tokens, we find that on IWSLT data, {\eurollm} and {\tower} models (the best at translation) are robust to random context and only degrade slightly in performance, aligning with previous observations of the minimal effect of context perturbation on translation performance. 
Additionally, those models (except {\eurollm} for \langpair{en}{fr}) show the least difference in {\GPR} performance between gold and random context setups among all LLMs. Robustness to total random context can be linked to lack of proper context utilization.
Although the TowerBlocks dataset used to finetune {\towerinst} models includes context-aware data (as per the dataset card\footnote{\href{https://huggingface.co/datasets/Unbabel/TowerBlocks-v0.2}{https://huggingface.co/datasets/Unbabel/TowerBlocks-v0.2}}), we hypothesize that general fine-tuning alone may not be sufficient for improving discourse phenomena performance. Explicit, context-aware fine-tuning might be required to address these challenges effectively.

\begin{figure*}[t]
\centering
\begin{subfigure}{0.49\textwidth}
    \includegraphics[width=\textwidth]{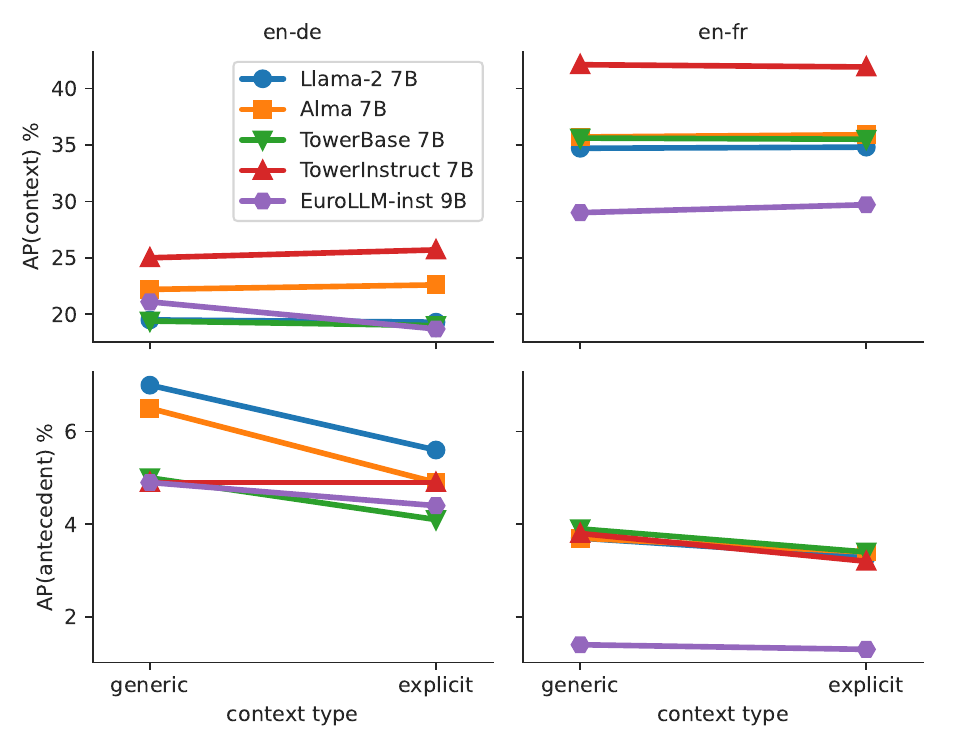}
    \caption{ALTI-Logit}
    \label{attr-alti}
\end{subfigure}
\hfill
\begin{subfigure}{0.49\textwidth}
    \includegraphics[width=\textwidth]{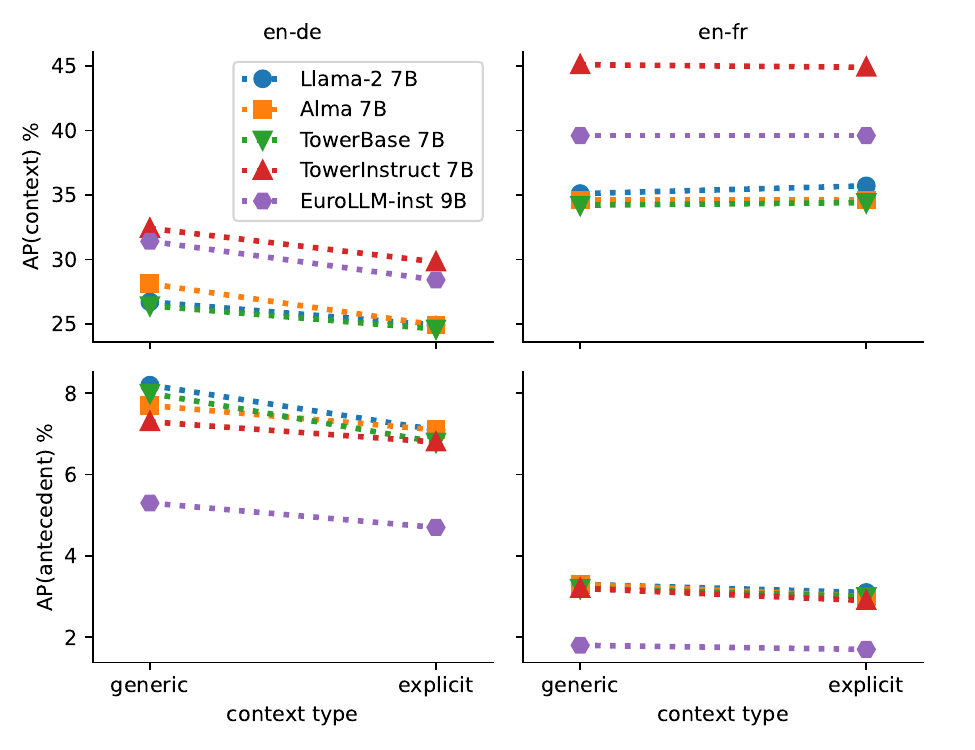}
    \caption{input-erasure}
    \label{attr-erase}
\end{subfigure}
\caption{Attribution percentages assigned to antecedent tokens (relevant context) and the entire context tokens when force-decoding the correct pronoun in ContraPro data. (a) shows results from ALTI-Logit and (b) displays results from input-erasure attribution methods.}
\label{attr-combined}
\end{figure*}

\subsection{Attribution Analysis}
We analyze models' internals to see how much the relevant context contributes to the outputs. \Cref{attr-alti,attr-erase} present attribution percentages of antecedent tokens (relevant context) as well as of the whole context using ALTI-Logit and input-erasure methods, respectively.

Looking at both attribution methods, we see that for {\eurollm} and {\towerinst} 7B (the best two models at translation among the 5 models tested) antecedent tokens have the lowest attribution percentage to the output. Even though for {\towerinst} 7B model, overall context tokens have the highest attribution percentage. This suggests that there is a need to explicitly finetune translation LLMs to focus on \textit{relevant context} at inference time.

However, unlike the larger differences in \textit{relevant context} and overall context attributions observed for encoder-decoder models by \citet{DBLP:conf/eacl/MohammedN24}, we find no striking differences or clear patterns between the contributions for LLMs. This might be due to the fact that the models have similar backbone structures.

\section{Related Work}\label{related_work}
\textbf{Context utilization assessment.} Works on assessing context utilization in machine translation include the work of \citet{DBLP:journals/corr/abs-2310-01188}, who build an end-to-end interpretability framework to quantify the plausibility of context-aware encoder-decoder machine translation models.
they leverage model internals to contrastively identify context-sensitive target tokens in generated texts and link them to contextual cues justifying their prediction. Using their approach, they were able to consistently detect context-sensitive tokens and their disambiguating rationales across several datasets, models and discourse phenomena. Inspired by this line of research, we evaluate context utilization of LLMs as a possible new paradigm for context-aware translation.\\

\noindent
\textbf{Perturbation and attribution analysis.} There are several works that used attribution and perturbation techniques to understand the inner workings of translation LLMs, mostly focusing on the in-context learning (ICL) paradigm --- a setup where LLMs "learn" to perform new tasks during inference by being provided with few task demonstrations in the input prompt. \Citet{DBLP:conf/emnlp/ZaranisGM24} use input attribution methods (ALTI) to examine context contributions in translation LLMs within the ICL paradigm. Their findings indicate that the source segments of few-shot examples contribute more significantly than their corresponding target segments, parallel-data fine-tuning alters contribution patterns, and context contributions exhibit a positional bias. \Citet{DBLP:journals/corr/abs-2310-15987} perturb in-domain translations to better understand their role in ICL. They perform asymmetric perturbation of source-target mappings and find that target perturbations has more negative effect on the translation performance compared to source perturbations. \Citet{DBLP:conf/naacl/ZhuLDXHKCL24} also perturb the in-context examples by providing unrelated task (summarization) examples and find that LLMs are not sensitive to the perturbation. 
Our work combines both interpretability techniques (perturbation and attribution methods) and focuses on context-aware translation task. 
\\

\noindent
\textbf{LLMs for document-level machine translation.} The line of research on adapting LLMs for document-level translation using techniques like LLMs fusion with translation models \citep{DBLP:conf/wmt/PetrickHPKN23}, finetuning LLMs on parallel document-level data \citep{DBLP:journals/corr/abs-2401-06468}, or a mix of sentence-level and document-level data \citep{DBLP:journals/corr/abs-2401-08088}, generally evaluates on translation metrics and discourse phenomenon accuracy. We complement such evaluations with a finer grained strategy that focuses on the role of context.

\section{Conclusion}\label{conclusion}
We use interpretability tools (perturbation and attribution techniques) to analyze LLMs' context-utilization in document-level translation. 
Our experiments suggest that multilingual pretraining and translation-specific finetuning of LLMs pushes state-of-the-art translation performance beyond encoder-decoder models. However, we 
highlight that looking at discourse phenomena performance, LLMs 
show room for improvement.
We argue that more care is needed before
adopting LLMs as the new standard for document-level translation, and 
more focused evaluation must be considered. 
Future research directions include enhancing context-aware translation capabilities of LLMs, potentially through explicit finetuning, and creating datasets with supporting-context annotations for other discourse phenomena to enable extending context-utilization analysis to those phenomena.

\section{Limitations}\label{limitations}
Even though some API-only LLMs (GPT-3.5 and GPT-4) show significant translation improvement compared to encoder-decoder document-level transformers and commercial translation systems 
\citep{DBLP:conf/emnlp/WangLJZY0T23}, our analysis approach relies on access to model internals in order to be able to compute attributions of input tokens. Thus, we only used open-source LLMs in our study. 

Based on the availability of datasets with context-dependent linguistic phenomena that include supporting context annotations (ContraPro), we experimented only on \langpair{en}{de} and \langpair{en}{fr}. These two languages belong to the same language family and we leave it to future work to experiment on general translation on other language families.

We chose well-established evaluation metrics in the literature to assess pronoun resolution accuracy. However, we acknowledge the limitations of those metrics. The contrastive metric ({\CPR}) is not aligned with the generative training objective of models and the generative metric (\GPR) misses cases where the model generates the correct pronoun in a different location in the sentence than the target location. 

Due to computational constraints, we were only able to perform the attribution analysis on a small set of models. We hope our work inspires more research into understanding the inner-workings of translation models in context utilization.

For all models except the transformer encoder-decoder model trained from scratch, we do not have details about their training data. This trend of releasing and building on models with secret training data is concerning because it makes fair evaluation impossible.

\section{Ethics Statement}\label{ethics}

Nowadays, machine translation is a widely adopted technology, sometimes in sensitive, high-risk settings. Even though we propose a fine-grained approach to assessing context utilization, and highlight its importance as we see that improvements in translation performance does not necessarily reflect in discourse phenomena performance, we still rely on automatic evaluation which is imperfect. For systems deployed in critical scenarios, we believe a nuanced case-by-case evaluation is always necessary.

\section*{Acknowledgements}
We would like to thank Evgenia Ilia, Bryan Eikema and Di Wu for their valuable comments and discussions about this work.

This work is part of the UTTER project, supported by the European Union's Horizon Europe research and innovation programme via grant agreement 101070631.
VN also acknowledges support from the Dutch Research Council (NWO) via VI.Veni.212.228.

\bibliography{anthology,custom}

\appendix

\section{Sustainability statement}
\label{sustainability}
Our experiments with 13B parameter models run in 95h on 2 GPUs NVIDIA A100 PCIe, and draw 81.69 kWh. 
Based in Netherlands, 
This has a carbon footprint of 30.58 kg CO2e, which is equivalent to 2.78 tree-years. For all other models, the experiments run in 502h on 1 GPU NVIDIA A100 PCIe, and draw 222.08 kWh. 
Based in Netherlands,
This has a carbon footprint of 83.13 kg CO2e, which is equivalent to 7.56 tree-years \citep{lannelongue2021green}.

\end{document}